# ALVIO: Adaptive Line and Point Feature-based Visual Inertial Odometry for Robust Localization in Indoor Environments

KwangYik Jung[1][0000-0003-0049-0567], YeEun Kim[2][0000-0002-5349-0520], HyunJun Lim[2][0000-0002-0199-8488], and Hyun Myung[2][0000-0002-0050-8627]

[1] Department of Civil and Environmental Engineering,
Korea Advanced Institute of Science and Technology, Daejeon, 34141, Korea
[2] School of Electrical Engineering, KI-AI, KI-R,
Korea Advanced Institute of Science and Technology, Daejeon, 34141, Korea
{ankh88324, yeeunk, tp02134, hmyung}@kaist.ac.kr
http://urobot.kaist.ac.kr

**Abstract.** Abstract—The amount of texture can be rich or deficient depending on the objects and the structures of the building. The conventional mono visual-initial navigation system (VINS)-based localization techniques perform well in environments where stable features are guaranteed. However, their performance is not assured in a changing indoor environment. As a solution to this, we propose Adaptive Line and point feature-based Visual Inertial Odometry (ALVIO) in this paper. ALVIO actively exploits the geometrical information of lines that exist in abundance in an indoor space. By using a strong line tracker and adaptive selection of feature-based tightly coupled optimization, it is possible to perform robust localization in a variable texture environment. The structural characteristics of ALVIO are as follows: First, the proposed optical flow-based line tracker performs robust line feature tracking and management. By using epipolar geometry and trigonometry, accurate 3D lines are recovered. These 3D lines are used to calculate the line reprojection error. Finally, with the sensitivity-analysis-based adaptive feature selection in the optimization process, we can estimate the pose robustly in various indoor environments. We validate the performance of our system on public datasets and compare it against other state-of the-art algorithms (S-MSKCF, VINS-Mono). In the proposed algorithm based on point and line feature selection, translation RMSE increased by 16.06% compared to VINS-Mono, while total optimization time decreased by up to 49.31%. Through this, we proved that it is a useful algorithm as a real-time pose estimation algorithm.

**Keywords:** Visual Inertial Odometry (VIO), Visual Inertial Navigation System (VINS), 3D line features, sensitivity analysis, adaptive feature selection, indoor localization.

# 1 INTRODUCTION AND RELATED WORK

## 1.1 Loosely Coupled and Tightly Coupled VIO

Visual Inertial Odometry (VIO) can be classified into loosely coupled and tightly coupled methods depending on how data from the inertial measurement unit (IMU) and visual information are combined. In the loosely coupled method, the pose estimations based on IMU increment and visual information are calculated separately and then merged into one pose later. On the other hand, the tightly coupled method directly combines the IMU and visual sensor information to calculate a jointly optimized estimate with only one estimator. Compared to loosely-coupled methods, tightly coupled methods are more accurate and robust. Therefore, in this paper, a tightly-coupled method that combines IMU and mono cameras visual information is considered.

The tightly coupled method is divided into the Extended Kalman Filter (EKF)-based method and the optimization-based method. A representative example of the EKF-based method is Multi-State Constraint Kalman Filter (MSCKF) [1]. MSCKF considers only the body pose without including the feature information in the state variable, while including the previous camera poses. Therefore, it is possible to accurately estimate the position with multi-state constraints for a single feature. Robust Visual Inertial Odometry (ROVIO) [2], [3] improves the consistency and computational speed of the system by using the rotation angle and distance information of landmarks. To improve the tracking performance between images, the intensity error of the image patch is directly used, and this error is included in the innovation term of the EKF update step.

The optimization-based method incorporates the residuals of the state variables and features into the cost function and optimizes the position by iterative calculations. Open Keyframe-based Visual-Inertial SLAM (OKVIS) [4] uses keyframe poses with keypoints in the image frame and optimizes them using IMU-based constraints. For this purpose, robust keypoint and keyframe selection algorithms and marginalization techniques for weight reduction have been proposed. VINS-Mono [5] has high accuracy because of the visual-inertial alignment, which is the fusion of the IMU and the point feature-based Structure from Motion (SfM) and the optimization algorithm for the IMU, point features, and marginalization factors in the sliding window structure.

Because all VIO algorithms mentioned above use only point features, performance is not guaranteed in indoor environments where visual information is not diverse.

## 1.2 Visual SLAM by Point and Line Features

Visual SLAM is a technique for estimating poses based on the visual information extracted from mono/stereo/depth cameras. It is classified into the direct method and indirect methods according to how the information on the image is processed. The direct method estimates the poses of the camera by directly using the intensity or gradient value of the image. On the other hand, the indirect method first extracts the representative features and matches them with the features of the other image and then infers the poses that reduce the geometric error. Although the indirect method requires a relatively large computational cost because of the feature matching and outlier rejection steps, many applications prefer this indirect method due to its accuracy and robustness.

The point feature of the indirect method is the most representative feature, and it is popularly utilized by the vision-based localization methods. However, the point is vulnerable to low texture environment, motion blur, and illuminance changes.

To overcome this problem, methods using additional line features, which have a large amount of structural information, are proposed by some researchers. To obtain stereoscopic 3D lines from a stereo camera, 3D line reprojection errors are used [6], and a tightly coupled monocular visual-inertial odometry system based on point and line features is devised [7]. Those works use the LSD binary descriptor-based line, which is inaccurate, and even the algorithm used in conjunction with the point does not consider the efficiency of the calculation.

In this paper, we propose an optical flow-based line tracker to improve the line prediction performance and reliability of sparse depth 3D line calculation.

### 1.3   Sensitivity Analysis and Feature Selection

One way to reduce the computational complexity of image feature-based localization methods is to selectively use the extracted image features [8]. The methods for choosing a visual feature with valid information include the entropy theory (using features that minimize entropy in pose estimation) [9], covariance ratio approach by evaluating the effects of features and selecting it in a covariance matrix [10], the Joseph's covariance matrix approach [11], etc.

In this paper, we propose the factors that affect the performance of each feature tracker for point and line feature selection. Then, the sensitivity analysis of the feature tracker is performed in terms of each factor. Sensitivity analysis [12] is the study of how the uncertainty in the output of a model or system can be divided and allocated to different input sources. By calculating the threshold of the input elements corresponding to the threshold of tracker performance and using it in feature selection, the number of feature residuals used in the optimization process is reduced and the complexity of the Jacobian matrix operation can also be reduced.

### 1.4   Contribution

This paper proposes a robust pose estimation algorithm for indoor environments, namely ALVIO. The contributions are as follows:

- ALVIO is a tightly coupled visual-inertial odometry in which lines and points are used complementarily.
- For accurate sparse depth 3D line calculation, optical flow-based line prediction and merging between the matched line and predicted line are performed.
- A sensitivity analysis of the feature tracker is performed for the optimization based on point and line features. It reduces the amount of computation in the optimization process by adaptively selecting valid point and line features based on sensitivity measures.

4- We evaluate the performance of the proposed algorithm through comparisons with the latest VIO algorithms using public datasets.

The rest of this paper is structured as follows: In Section 2, we describe how to apply the line feature to the optimization based VIO. The optical flow-based sparse depth 3D line tracking method is described in detail. In Section 3, we explain the adaptive feature selection process through an in-depth sensitivity analysis of the point and line feature trackers. Finally, the experimental results for several datasets are shown in Section 4, while Section 5 concludes the paper.

## 2    VIO WITH LINE FEATURES

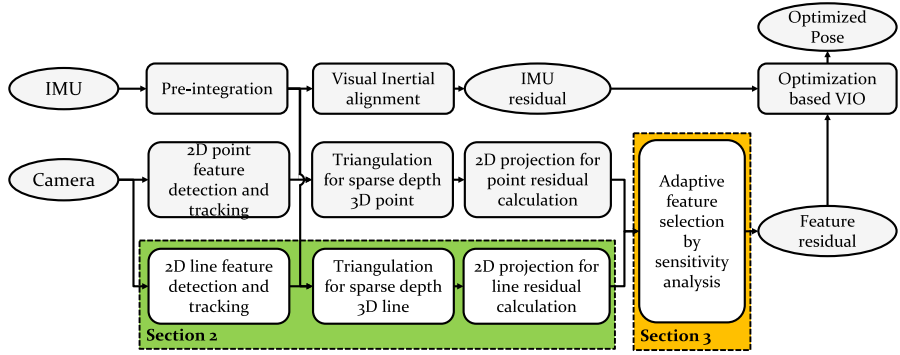

**Fig. 1.** Overall block diagram of ALVIO. ALVIO is an extension algorithm that considers the line feature in VIO to perform robust localization in indoor environment. The novelties of ALVIO include optical flow-based line tracker (Section 2) and the sensitivity-analysis-based adaptive feature selection (Section 3) algorithm.

The overall block diagram of the VIO algorithm considering the line features is shown in Fig. 1. The basic structure of our algorithm follows VINS-Mono [5] and PL-VIO [7]. VINS-Mono, which is a sliding window-based algorithm using two measurements (IMU and point feature), performs pose estimation by minimizing the sum of prior and all measurement residuals. ALVIO is an extension of the VINS-Mono algorithm, which incorporates the line features in VINS-Mono algorithm to perform robust localization in indoor environments. The novelties of ALVIO include the optical flow-based line tracker (Section 2.2) and the sensitivity-analysis-based adaptive feature selection (Section 3) algorithm.

### 2.1    Definition of State Variables

The system state variables we have defined are as follows:

$$\mathcal{X} = [x_0, x_1, \cdots, x_{k-1}, \lambda_0, \lambda_1, \cdots, \lambda_{m-1}, o_0, o_1, \cdots, o_{n-1}] \quad (1)$$

$$x_i = [p_{b_i}^w, q_{b_i}^w, v_{b_i}^w, b_a^{b_i}, b_g^{b_i}] \quad (2)$$

$$o_j = [\psi_j, \phi_j] \quad (3)$$

where $\mathcal{X}$, $k$, $m$, and $n$ are the full state variable, the number of sliding windows, the number of point features, and the number of line features, respectively. Here, $x_i$ is the IMU state of the $i$-th sliding window, including the position $p_{b_i}^w$, orientation $q_{b_i}^w$, velocity $v_{b_i}^w$, and bias of the accelerometer $b_a^{b_i}$, bias of the gyroscope $b_g^{b_i}$. Also, $\lambda_i$ is the inverse depth relative to the first image frame at which a point is first detected, which follows the VINS-Mono representation. For the sparse depth 3D line features $o_j$ (composed with the 3-DoF Euler angle ($\psi_j$) and the 1-DoF rotation angle ($\phi_j$)), we use the 4-DoF orthonormal representation [13], which is efficient because it reduces the number of parameters compared to the existing 3D line representation (one 3D point with a direction vector that passes through the point).

## 2.2 Optical Flow-based Robust Line Tracker

To use the line features in VIO, the line features must be tracked seamlessly in the sliding window. However, it is difficult to track line features when using Line Segment Detector (LSD) [14]. LSD is the method of segmenting similar angle regions by calculating a level-line field for gradient changes in gray images. Following this principle leads to computational inefficiencies in two situations: when the long line extracted from the previous frame is divided into short lines in the current frame and when a different line ID is given because a part of the line extracted from the previous frame is out of the current frame.

To solve this problem, we propose an optical flow-based line feature tracker as shown in Fig. 2 There are four types of lines when comparing the LSD-based line extracted from the current frame ($f_k$) with that extracted from the previous frame ($f_{k-1}$).

- Type1: Lines that retain both end points in $f_{k-1}$ and $f_k$.
- Type2: Newly extracted line from $f_k$.
- Type3: The line extracted from $f_{k-1}$ is divided into multiple lines in $f_k$.
- Type4: A line with one of the end points from $f_{k-1}$ is out of the range of the current image frame $f_k$.

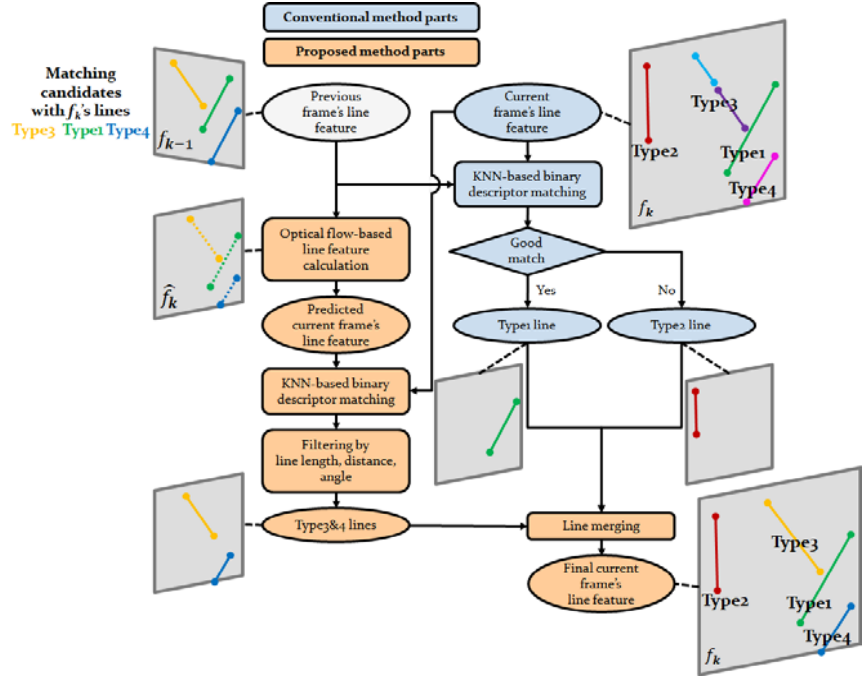

**Fig. 2.** Block diagram illustrating the full pipeline of the proposed optical flow-based line tracker with an example image for each step**.**

The existing line tracking method performs LSD binary descriptor matching based on k-Nearest Neighbors (KNN) [15] for $f_k$ lines and $f_{k-1}$ lines. Through this process, the Type1 line is extracted as a good matching element. The Type2 line is not matched but is a newly extracted line and can be used for the line tracker of $f_{k+1}$. If previous processes work only, the Type3 and 4 lines are not used nor classified as new lines, which reduces the feature tracker performance and computational efficiency.

Our algorithm calculates the optical flow at the end points of the $f_{k-1}$ line. At this time, if the end points based on the optical flow exceeds the image frame region, the slope of the line is calculated and updated to the end point coordinates in the image frame region. Through this, we can estimate the end point position in $\hat{f}_k$ and perform KNN matching with actual $f_k$ line endpoints. Here, the $\hat{f}_k$ is the estimated image frame based on optical flow.

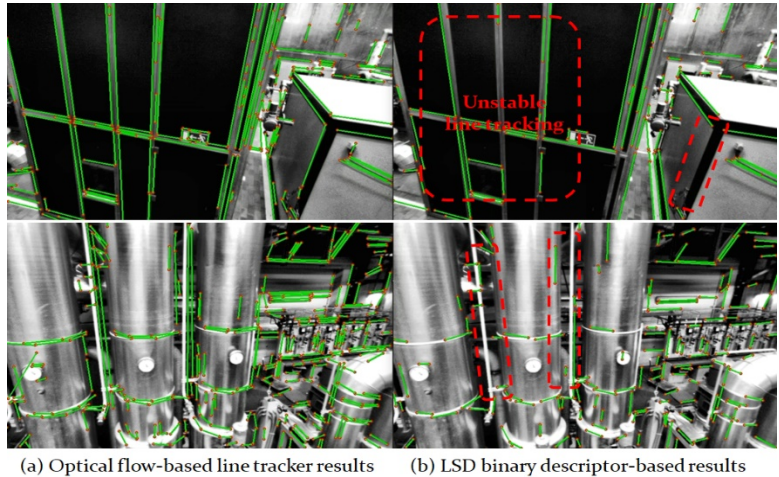

(a) Optical flow-based line tracker results    (b) LSD binary descriptor-based results

**Fig. 3.** Optical flow-based line feature tracking result.

Subsequently, filtering is performed according to the line length, distance between line center coordinates, and line angle. As a result, two lines divided in the $k$-th frame are refined into a single line of Type3, and the same ID of the Type4 line is maintained. Fig. 3 shows the exemplary result, which demonstrates that the line tracking works robustly in successive frames with this algorithm.

## 3   SENSITIVITY ANALYSIS BASED ADAPTIVE FEATURE SELECTION

The system proposed in this paper requires a large amount of computation because it uses the point and line features together. Therefore, in order to reduce the computational cost while maintaining the performance, we analyzed the various factors that affect the performance of the feature tracker through sensitivity analysis. As shown in Fig. 4, the sensitivity analysis was performed on two main categories. Firstly, the factor affecting the importance of individual features was examined. Then, the different environmental conditions that can affect the relative importance between point and line features were evaluated. Therefore, point and line features could be adaptively selected and fused with different weights based on the analysis.

### 3.1   Sensitivity Analysis

Sensitivity analysis is one of the methods used to examine the outputs of the model by substituting all possible input values the model can take. It is especially useful when the parameters that can affect the model are uncertain. It can be said that if one 3D feature is continuously tracked in successive frames, the feature would contain considerable information about the correlation of the 3D pose and constraints between frames. Therefore, in this paper, the performance of the feature tracker is defined by how long the feature has been tracked in the sliding window, and it is analyzed using sensitivity analysis.

### 3.2 Sensitivity Analysis on Feature Tracker for Feature Selection

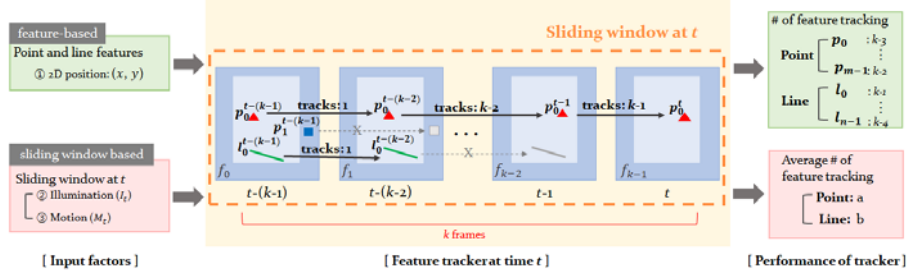

**Fig. 4.** Sensitivity analysis diagram. The analysis was performed on two main categories: (1) The factor affecting the importance of individual features (feature-based factor). The factor considered for this is the location in the 2D image plane. From a given input factor, the feature tracker performance is evaluated by the number of tracking of each feature. (2) The factors affecting the relative importance between point and line features (sliding-window-based factors). Average illumination ($I_t$) and motion ($M_t$) of the sliding window at time $t$ were used as input factors. Likewise, the performance of tracker is validated by the average number of tracked features in the first frame of the sliding window. In the box at the middle, feature tracking in the sliding window is illustrated.

To analyze the characteristic of individual visual features affecting the performance of the feature tracker, the factor we consider is the pixel coordinates of each feature in the 2D image plane.

To numerically analyze the performance of the tracker on individual point and line features, we used the number of times each feature was tracked in the sliding window as the evaluation criteria. For a given input factor corresponding to the factor written in the green box in the upper left corner in Fig. 4, we checked how long the point and line features in the first frame of the sliding window were tracked. In Fig. 4, $p_m^t$ refers to the position of the $m$-th point feature in the 2D image plane at time $t$, and similarly $l_n^t$ refers to the position of the $n$-th line feature in the 2D image plane at time $t$. Also, $f_k$ represents the $k$-th frame in the sliding window. It can be assumed that the feature located at the edge of the image plane is less likely to be tracked according to the camera movement. Based on this assumption, we checked the effect of the location of the feature in the 2D plane on the performance of the trackers.

For this, we plotted how many times the features in $f_0$ were tracked, and fitted them to a polynomial surface as shown in Fig. 5 (a) and (b). The tracker performance is degraded for both X and Y coordinates, when the features are located at the edges (ends of each graph) of the image. Besides, for the Y coordinate, the graph is not symmetric about the middle value (unlike that for the X coordinate), and the performance of the tracker tends to become worse as the value of the Y coordinate increases. This might be because many structures or objects lying on the floor are close to camera, and line features, which contain structural information in the upper part of the image, are usually distant from the camera as the characteristics of the indoor environment. On the other hand, for the X coordinate, the tracker performance is equally degraded for the edge region of the image plane.

Based on the analysis results above, it can be concluded that the position in the 2D image of each feature affects the tracker performance and the feature that is located at the center of the image plane is more important. By using this, we set the threshold value for feature selection, resulting in the reduced computation, while achieving similar performance compared with the-state-of-the-art algorithms like Table 2.

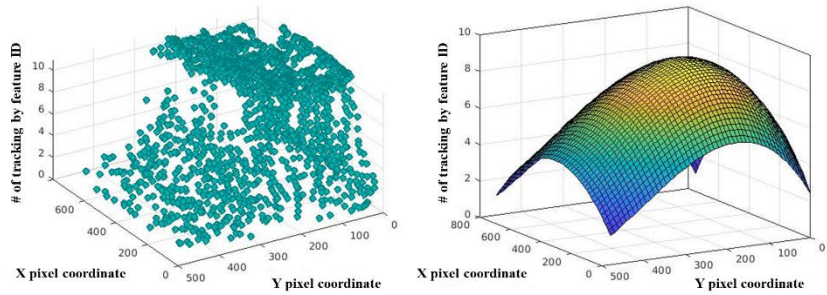

**Fig. 5.** Sensitivity analysis w.r.t the location in the 2D image plane. The obtained results on 2D location were visualized. (left) Obtained raw data. (right) Fitted polynomial surface. The colors used for polynomial surface indicate the length of feature tracking (yellow: big value, blue: small value).

### 3.3  Sensitivity Analysis on Feature Tracker for Feature Fusion

Similarly, for understanding the environmental conditions affecting the performance of the feature tracker, two factors are considered: 1) average illumination of the image frames in the sliding window and 2) camera motion between the first and last frames in the sliding window.

As shown in Fig. 4, to evaluate the tracker performance under environmental conditions that correspond to the pink box at the bottom left, the number of tracking of the features have been tracked in the sliding window is used for the performance analysis. However, to analyze the performance of the sliding window at time $t$, not a single feature, according to the given condition, the tracker performance is calculated by the average value of the number of tracking of each feature (pink box at the bottom right) in the sliding window.

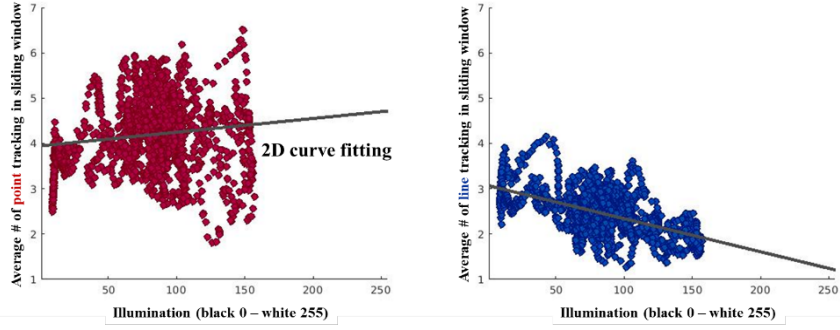

**Fig. 6.** Sensitivity analysis w.r.t the illumination. The average number of tracking of (left) point features and (right) line features in the first frame according to average illumination in the sliding window.

Fig. 6 shows the average number of tracking of the point and line features according to the illumination ($I_t$) value of the images through frames $f_0$ to $f_{k-1}$ in the sliding window at time $t$. For calculating the illumination, the average pixel value in a single frame is used. As shown in Fig. 6, the line features tend to be more robust in a low illumination environment unlike the point tracker

The rationale for this is that when the optical flow method is used for point tracking, only the changes of the adjacent pixels are considered, while the line feature calculates the changes over a wide range, allowing robust tracking even in a low illumination environment.

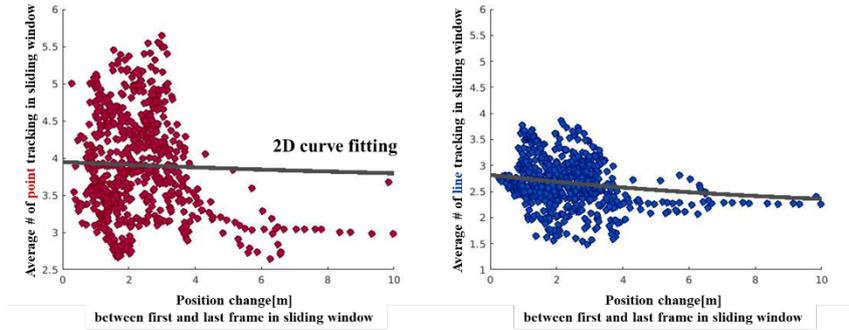

**Fig. 7.** Sensitivity analysis w.r.t the motion. The average number of (left) point tracking and (right) line tracking according to the motion between the first and last frames in the sliding window.

Lastly, tracker performance was analyzed according to the amount of the position change (motion) between the first and last frames of the sliding window at time $t$. Here, the motion ($M_t$) for sliding window at time $t$ is defined as the sum of the absolute values of the positional difference between the first and last frames in the sliding window at time $t$. Fig. 7 shows the average number of tracking of the point and line features in the sliding window against the amount of motion between the first and last frames in the sliding window. The tracker performance tends to decrease slightly for both points and lines as the change in position increases.

As a result of the performance analysis of the tracker under the varying environmental conditions, it was found that the point and line features had different tracker performances depending on the average illumination of the image frames and the amount of the position change in the sliding window. Therefore, we can improve the estimation accuracy by adaptively weighting the point and line features depending on the situation as follows:

$$w_p = \frac{M_p}{\alpha M_l + M_p}, w_l = \frac{\alpha M_l}{\alpha M_l + M_p} \quad (4)$$

where

$$\alpha = \begin{cases} \alpha_1, & \text{if } M_l > M_p \\ \alpha_2, & \text{if } M_l \approx M_p \\ \alpha_3, & \text{if } M_l < M_p \end{cases}$$

where $M_p$ and $M_l$ represent each feature's number of tracking and $w_p$ and $w_l$ represent the relative weight for point and line features, respectively. Because the number of feature tracking $M_p$ is much larger than $M_l$, we use the adaptive gain $\alpha$ which is multiplied with $M_l$ according to the situation.

## 4    EXPERIMENTAL RESULTS

### 4.1    Evaluation of Adaptive Feature Selection

As shown in Fig. 8, we analyzed the number of point and line features used for calculating the residual and for optimization process. By comparing the tracking capability of the point tracker and line tracker, we calculate the relative weights over time. Through the experiments, we get the value of $\alpha$ in (4) roughly. $\alpha_1 = 5.0$, $\alpha_2 = 2.2$, and $\alpha_3 = 1.0$. We can see that the point tracker and line tracker perform similarly or complementarily in different situations. Also, the line tracker is more robust compared to the point tracker under weak illumination. In this graph, the point feature is detected more than the line feature, but it has a big change in the rate of increase and decrease. On the other hand, the line has undergone preprocessing to maintain tracking information between consecutive frames based on the optical flow. Therefore, the tracker performance is less sensitive than point features. In other words, even if the number of line features is small compared to that of point features, it can be confirmed that it operates as a more robust feature in the indoor environment.

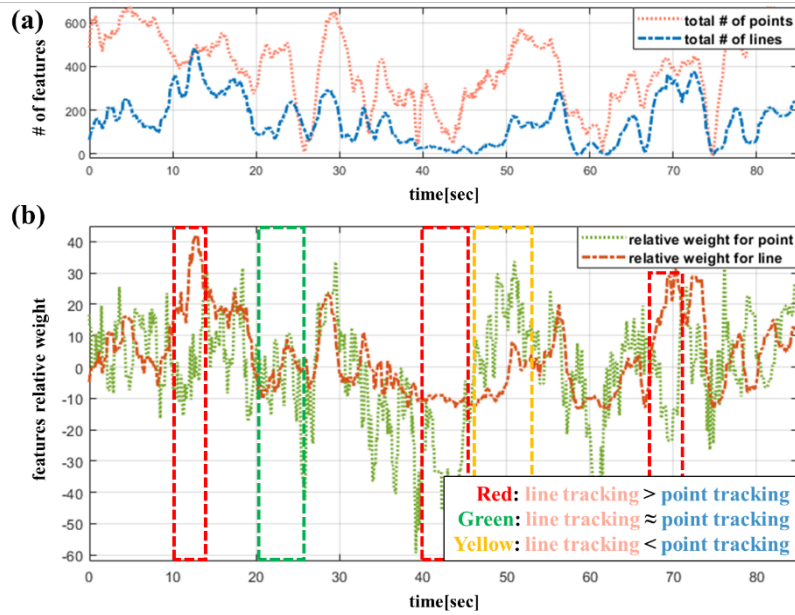

**Fig. 8.** Evaluation graph about sensitivity-analysis-based feature selection and weighting algorithms. (a) Each feature's adaptive feature selection results. (b) The plot of relative weights over time. The red, green, and yellow dotted blocks represent the performance comparison of the line tracker and point tracker.

### 4.2 Comparison with State-of-the-art Algorithms

For evaluating the proposed ALVIO system, an experiment was conducted using a public dataset and analyzed the factors that have an effect on the system performance. Firstly, our proposed algorithm was compared with other state-of-the-art algorithms on the public dataset to perform a numerical analysis on the performance of our system. We also tested and compared the performance when the adaptive feature selection algorithm was applied.

All the experiments were conducted on a computer equipped with an Intel Core i7-4710MQ CPU with 2.50 GHz, 8 GB RAM, and ROS Kinetic (ROS). As the dataset, we used EuRoC MAV dataset [16], which contains stereo images (20 FPS) collected by a micro-aerial vehicle and synchronized IMU (200 Hz) data. The ground truth data captured via visual motion capture is also provided. Since our algorithm is based on a monocular camera, we used only the left camera's images during the experiment. For the extrinsic and intrinsic parameters, we applied the values provided by the dataset.

Table 1 shows the comparison result with other state- of-the-art systems: (1) VINS-Mono, which is a mono VIO algorithm based on visual point features, (2) S-MSCFK, the filter based VIO expanding the Multi-State Constraint Kalman Filter (MSCKF) algorithm to stereo images. Since S-MSCKF does not have the loop closure process, we measured the accuracy without the loop closure for a fair comparison.

**Table 1.** Comparison with other state-of-the-art visual odometry algorithms using EuRoC datasets. For a fair comparison, the loop closure procedure is excluded.

| Algorithms | RMSE[m] | | | |
|---|---|---|---|---|
| | Sequence | | | |
| | MH_01 | MH_02 | MH_03 | MH_05 |
| VINS-Mono [5] | 0.15602 | 0.17841 | 0.19487 | 0.30234 |
| S-MSCKF [17] | - | **0.15213** | 0.28959 | 0.29312 |
| Proposed | **0.10637** | 0.17791 | **0.19005** | **0.28046** |

We evaluated the algorithms on different datasets: Machine Hall (MH) 01 easy, MH 02 easy, MH 03 medium, and MH 05 difficult. Table 1 shows the root mean square error (RMSE) of each algorithm on the different sequences of EuRoC dataset. The result shows that ALVIO outperforms the other algorithms in most cases. In indoor environments such as Machine Hall, where artificial structures are abundant, it was confirmed that the proposed algorithm using the line feature operates more robustly than the point only algorithms like VINS-Mono and S-MSCKF.

**Table 2.** Sensitivity analysis-based feature selection results. Comparison analysis of optimization time and RMSE according to each selection of points and lines in EuRoC dataset MH_05. (x) means without selection and (o) means feature selection is performed.

| | Point (x) Line (x) Reference | Point (x) Line (o) | Point (o) Line (x) | Point (o) Line (o) |
|---|---|---|---|---|
| The number of point features | 995 | 1009 | 243 | 263 |
| The number of line features | 294 | 122 | 288 | 130 |
| Total optimization time[msec] | 47.051 | 41.404 (reduced 12.00%) | 28.081 (reduced 40.31%) | 23.849 (reduced 49.31%) |
| Translation RMSE[m] | 0.3100 | 0.3232 (increased 4.26%) | 0.3413 (increased 10.09%) | 0.3698 (increased 16.06%) |

The number of features in the second column of Table 2 refers to the average number of features per frame extracted from the MH_05 dataset. This is considered as a reference. In contrast, the third column in Table 2 refers to the case where selection is applied only for line features. The fourth column means when selection is applied only for point features. In the last column, the optimization time was reduced by 49.31% whereas the RMSE was increased by only 16.06% as a result of all feature selection, in contrast to the nonselective situation. Feature selection reduces the number of 3D features that are the target of reprojection error calculations. According to the reference (the first column in Table 2), in general, point features compared to line features are used for pose estimation. Therefore, it can be seen that the effect of point selection has a great effect on increasing RMSE and decreasing optimization time.

## 5    CONCLUSIONS

In the proposed ALVIO, the line and point features play a complementary role in the optimization-based visual-inertial odometry. In ALVIO, optical flow-based line tracking was performed, and sparse depth 3D lines based on epipolar geometry and trigonometry between multiple frames were calculated. By performing sensitivity analysis on the feature tracker and optimizing it with the residuals of the adaptively selected features, the amount of computation was reduced while retaining the accuracy. We compared and verified the performance of the proposed algorithm by comparing it with the latest VIO algorithms using public datasets. As a feature-based algorithm, this paper confirms that line features are useful in indoor environments with many artificial structures. Furthermore, it is possible to realize the necessity of an algorithm that selects only features useful for pose estimation among abundant features.

We will improve ALVIO and turn it into a mobile algorithm. Recently, Samsung Note 10 was installed in an unmanned vehicle to provide Virtual Reality (VR) video information [18]. This means that the quality of vision and inertial navigation sensor data from mobile devices are approaching to that of expensive sensor systems designed for precise localization. If ALVIO is optimized, minimized, and embedded to operate with mobile sensor data, it is expected to show excellent scalability by being used for pose estimation in applications such as AR (augmented reality) in mobile devices, and payload-sensitive small drones and robots that can move in small spaces.

## 6    Acknowledgement

This work was supported by the Defense Challengeable Future Technology Program of Agency for Defense Development, Republic of Korea. The students are supported by Korea Ministry of Land, Infrastructure and Transport (MOLIT) as "Innovative Talent Education Program for Smart City" and BK21 FOUR.